\documentclass[conference]{IEEEtran}
\IEEEoverridecommandlockouts
\usepackage{amsmath,amssymb,amsfonts}
\usepackage{algorithmic}
\usepackage{graphicx}
\usepackage{textcomp}
\usepackage{xcolor}
\usepackage[maxbibnames=2,natbib=true]{biblatex}
\bibliography{reference}
\usepackage{float}
\usepackage{hyperref}
\usepackage{multirow}
\usepackage{booktabs}

\usepackage{fancyhdr}
\def\BibTeX{{\rm B\kern-.05em{\sc i\kern-.025em b}\kern-.08em
    T\kern-.1667em\lower.7ex\hbox{E}\kern-.125emX}}
\begin{document}

\title{Ensemble Learning to Assess Dynamics of Affective Experience Ratings and Physiological Change\\
\thanks{All authors contributed equally to this work.}
}

\author{\IEEEauthorblockN{Felix Dollack\textsuperscript{1}, Kiyoshi Kiyokawa\textsuperscript{1}, Huakun Liu\textsuperscript{1}, Monica Perusquia-Hernandez\textsuperscript{1}, \\
Chirag Raman\textsuperscript{2}, Hideaki Uchiyama\textsuperscript{1}, Xin Wei\textsuperscript{1}}
\IEEEauthorblockA{
\textit{\textsuperscript{1}Nara Institute of Science and Technology}\\
\textit{\textsuperscript{2}Delft University of Technology}\\
\{felix.d, kiyo, liu.huakun.li0, m.perusquia, hideaki.uchiyama, wei.xin.wy0\}@is.naist.jp, c.a.raman@tudelft.nl}
}

\maketitle
\thispagestyle{fancy}
\begin{abstract}
The congruence between affective experiences and physiological changes has been a debated topic for centuries. Recent technological advances in measurement and data analysis provide hope to solve this epic challenge. Open science and open data practices, together with data analysis challenges open to the academic community, are also promising tools for solving this problem. In this entry to the Emotion Physiology and Experience Collaboration (EPiC) challenge, we propose a data analysis solution that combines theoretical assumptions with data-driven methodologies. We used feature engineering and ensemble selection. Each predictor was trained on subsets of the training data that would maximize the information available for training. Late fusion was used with an averaging step. We chose to average considering a ``wisdom of crowds'' strategy. This strategy yielded an overall RMSE of 1.19 in the test set. Future work should carefully explore if our assumptions are correct and the potential of weighted fusion.
\end{abstract}


\section{Introduction}
Understanding human emotion is instrumental for applications in mental healthcare, education, and communication~\cite{calvoOxfordHandbookAffective2015}. These applications aim to automatically assess and generate affective cues by relying on an assumed relationship between affective experience and physiological changes. However, the debate on the precedence of body changes or subjective experience started in the previous century and remains current~\cite{cannonJamesLangeTheoryEmotions1987b}. Recent research has discussed whether demand characteristics affect bias in our understanding of the relationship between facial expression and affective experience~\cite{colesFactArtifactDemand2022a,barrettEmotionalExpressionsReconsidered2019a}; and explored the relationship between dynamic Autonomic Nervous System responses and affective experiences~\cite{gollandStudyingDynamicsAutonomic2014,pasquiniDynamicAutonomicNervous2023}.
Physiological sensing technologies have been popular in studying the physiological changes correlating with affective experiences~\cite{cacioppoInferringPsychologicalSignificance1990a,gunesAutomaticFacialExpression2016}. 
Each physiological measurement type gives a different piece of information regarding the functioning of the sympathetic and parasympathetic nervous systems~\cite{baltersCapturingEmotionReactivity2017}, leading to a popular multidimensional dataset collection. Traditional data analysis techniques require extensive knowledge about physiology characteristics, signal processing, and domain knowledge in affective sciences. This domain knowledge gave birth to hand-crafted feature engineering that improves data interpretability and reduces the number of comparisons to be made when analyzing the data. 
Recent advances in Machine Learning~(ML) and data-driven analyses have brought a new perspective. Purely data-driven analyses with end-to-end automated processing have become popular~\cite{liDeepFacialExpression2022}. In end-to-end approaches, a machine learning network learns an intermediate representation of the input, thereby reducing manual work, and potentially enhancing the results~\cite{kerenEndtoendLearningDimensional2017}. However, the evidence does not always support this claim. A previous study showed that convolutional and recurrent neural networks yielded better results than other state-of-the-art methods~\cite{kerenEndtoendLearningDimensional2017}. Another study used a deep-learning approach to estimate momentary emotional states from multi-modal physiological data; and reported a higher correlation than traditional methods. Still, their mean absolute error (MAE) was higher (a lower MAE is better)~\cite{hssayeni_multi-modal_2021}. Finally, another study found that end-to-end processes are suitable for predicting stress states with abrupt changes, but not as good when assessing subtle affective states like enjoyment~\cite{ditch_features20}. Hence, end-to-end learning only provides a marginal improvement over feature engineering for physiological signal-based affect recognition. This is different from camera-based recognition, where performance is radically improved. One possible explanation is the limited amount of physiological data publicly available. Therefore, public data sets and multi-laboratory collaborations are necessary to assess the effectiveness of different training methods and cross-validation strategies.

The EPiC challenge aims to overcome the limitations in data availability and motivates researchers to work on the affect-embodiment coherence problem. 
Our team used theory-driven analysis and data engineering techniques to address the EPiC challenge. We opted for feature engineering and ensemble learning for our final submission. 
Also, we report an exploratory analysis validating our assumptions for the challenge submission.

\section{Related Work}
\label{sec:relatedwork}
ML has been used to model emotion recognition mechanisms from data following the public release of benchmark databases.
The basic procedure of classical ML-based methods consists of four steps: physiological signal collection while eliciting participants' emotions, feature extraction from the signals, training a classification model with the features, and emotion recognition based on the trained model~\cite{liDeepFacialExpression2022}.
Research issues include the design of discriminative features and the selection of the optimal classification technique.
For instance, hypothesis testing is performed over some features followed by a predictive model that makes feature selection to see if the tested features were still relevant when all features were considered together~\cite{vailVisualAttentionSchizophrenia2017}.
It has been suggested that group synchrony improved arousal and valence classification~\cite{bota2023}.
Electrocardiograms (ECG) and Electrodermal Activity (EDA) have been used as tabular data with  AutoGluon-Tabular to arousal and valence across individuals and datasets~\cite{chhan_evaluation_nodate} with similar accuracy (around $56-62\%$ respectively) to previous works. Nevertheless, the subject-independent classification remains only slightly above chance level. 

A crucial challenge surrounding continuous-time annotation of emotions is the lag between observed features and the reported emotion measures~\cite{schmittBorderAcousticsLinguistics2016, ringeval2014prediction}. 
This lag arises from the time the rater requires to provide feedback about the experienced emotion.
Such a temporal misalignment between features and labels has consequences for ML methods.
Consequently, several compensation techniques have been investigated~\cite{nicolaou2010automatic, nicolle2012robust, mariooryad2013analysis, mariooryad2014correcting}.
These methods involve estimating the reaction lag from the data, by maximizing the correlation coefficient~\cite{nicolaou2010automatic, nicolle2012robust, mariooryad2013analysis, schmittBorderAcousticsLinguistics2016} or the mutual information between the multimodal features and emotional ratings~\cite{mariooryad2014correcting}. Others have used a recurrent neural network to handle the asynchronous dependencies~\cite{ringeval2014prediction}. 

Deep learning~(DL) has also been used in affective computing.
Handcrafted feature design is not always necessary in DL~\cite{liDeepFacialExpression2022}, also, less modalities seem to be required to achieve equal performance. For example,
\citet{hssayeni_multi-modal_2021} presented a DL approach to estimate momentary emotional states from multi-modal physiological data. Used modalities included respiration, ECG, electromyography (EMG), EDA, and acceleration. The best emotion classification was achieved by a traditional method with 79\% F1-score when all four physiological modalities were used. In contrast, using only two modalities, DL achieved 78\% F1-score.
Furthermore, there are several fusion strategies for multi-modal data: feature-level fusion, decision-level fusion, model-level fusion, and hybrid-level fusion. Late fusion by averaging class probabilities has performed well in the past~\cite{hssayeni_multi-modal_2021}. 

In the case of continuous detection of valence and arousal, a previous work obtained 0.43 and 0.59 RMSE for valence and arousal, respectively~\cite{siirtolaPredictingEmotionBiosignals2023}, in the WESAD dataset. When dividing the continuous annotation into binary valence-arousal categories (high-low), another group of researchers reported subject-independent accuracy of 76.37\% and 74.03\% for valence and arousal, respectively~\cite{zhangCorrNetFineGrainedEmotion2021b}, on the Continuously Annotated Signals of Emotion (CASE) dataset~\cite{sharmaDatasetContinuousAffect2019a}.


\section{Challenge corpora}
\begin{figure}[t]   
\centering
\includegraphics[width=0.7\columnwidth]{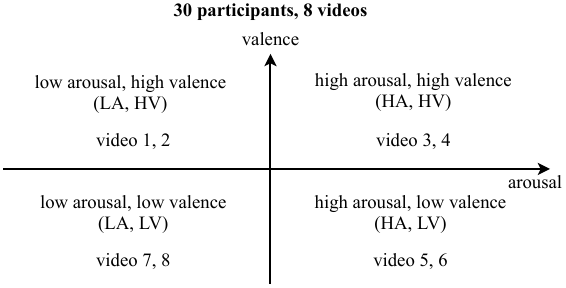}
    \caption{Diagram depicting the dataset structure utilized in this competition.}
    \label{fig:data_scription}
\end{figure}

The challenge corpora is an open dataset that collected six physiological signals while $30$ participants ($15$ female, age range: $22-37$ years) watched eight videos~\cite{sharmaDatasetContinuousAffect2019a}. The videos aimed to elicit a range of emotions and were rated with continuous self-reported valence and arousal in the range of 0.5 to 9.5 using a joystick. Visual feedback was provided using the Self-Assessment Manikin~\cite{bradleyMeasuringEmotionSelfassessment1994b}. Two videos were chosen per quadrant in the valence-arousal space, often called affect grid~\cite{russellAffectGridSingleItem}, and were presented in pseudo-random order to the participants. \autoref{fig:data_scription} shows the video distribution. The physiological sensing was logged at 1000~Hz, and the continuous annotation was done at 20~Hz. The physiological sensors included are:

\begin{enumerate}
    \item Cardiac activity as measured from Electrocardiography (ECG) and Photoplethysmography (BVP).
    \item Muscle activity (EMG) recorded from three muscles: the corrugator supercilii (emg\_coru), the zygomaticus major (emg\_zygo), and the trapezius (emg\_trap). 
    \item Electrodermal activity (EDA) measured from the non-dominant hand.
    \item Respiration (RSP) recorded from the chest.
    \item Skin Temperature (SKT) recorded from the little finger of the non-dominant hand. 
\end{enumerate}

For the EPiC Challenge, the Dataset was arranged in four scenarios to test four assumptions about the relationship between affective experiences and embodied cues. Each scenario is divided into training and test sets, as prepared by the organizers~\footnote{See: https://github.com/Emognition/EPiC-2023-competition}.

\subsection{Across-time scenario}
This scenario evaluates subject-dependent and affective context-dependent model performance.
The model is trained and tested over different durations of one data file (sub\_vid).
In this scenario, each of the 240 data files, that is, 30 participants watching eight videos, is divided into training and test parts based on the time series.
For each data file, the earlier part is the training data, and the latter is the test data.
The training and test data are not consecutive but are spaced by an unknown length of time.
The length of the training data ranged from 48~s to 127~s depending on the size of each video, with an average of 88~s.
All test data files were 50~s in length.

\subsection{Across-subject scenario}
This scenario evaluates subject-independent model performance.
The model is trained on data from some participants and tested on data from another set of different, unseen, participants to verify the model's generalization ability to new people.
In this scenario, the data of 30 participants were divided into five groups.
Each group contains six participants watching eight videos for a total of 48 data files.
This scenario consisted of five folds to use the cross-validation strategy.
In each fold, the data of four groups of participants were set as the training data (192 data files in total).
The 48 data files from the remaining six participants were set as the test data.
The length of the training data files ranged from 50~s to 128~s, with an average of 90~s. All test data files lasted 50~s.

\subsection{Across-elicitor scenario}
This scenario evaluates affective context-independent model performance.
The model is trained on data from several affective contexts and then tested on data from a different affective context.
Each affective context represents one quadrant in the valence-arousal affect grid. There were two elicitors (i.e., videos) per affective context.
This verifies whether the model can infer from the physiological signals triggered by one affective context to the physiological features triggered by another affectivity.
In this scenario, the data from eight videos were divided into four groups, according to the video's affective context. This resulted in four categories: low valence, high arousal; high valence, high arousal; high valence, low arousal; and low valence, low arousal.
Each group contains 60 data files, which corresponds to 30 participants with two videos each. 
The two videos in one group are considered to trigger the same type of affectivity.
By adopting a cross-validation strategy, this scenario contains four folds.
In each fold, three groups, totaling 180 data files, were chosen as the training data for the challenge.
The remaining two videos for a total of 60 data were chosen as the test data.

\subsection{Across-version scenario}
This scenario evaluates affective context-dependent model performance.
The model is trained on data from a specific affective state instantiation and then tested on the other version of the same affective contexts to verify the model's generalizability across similar affective contexts.
The data from eight videos were divided into two groups.
Each group contains 30 users watching four video types covering all quadrants in the affect grid, for a total of 120 data files.
This scenario consists of two folds.
In each fold, one group is the training data, while the other is the test data.

\section{Tackling the challenge}

\begin{figure}
    \centering
    \includegraphics[width=\columnwidth]{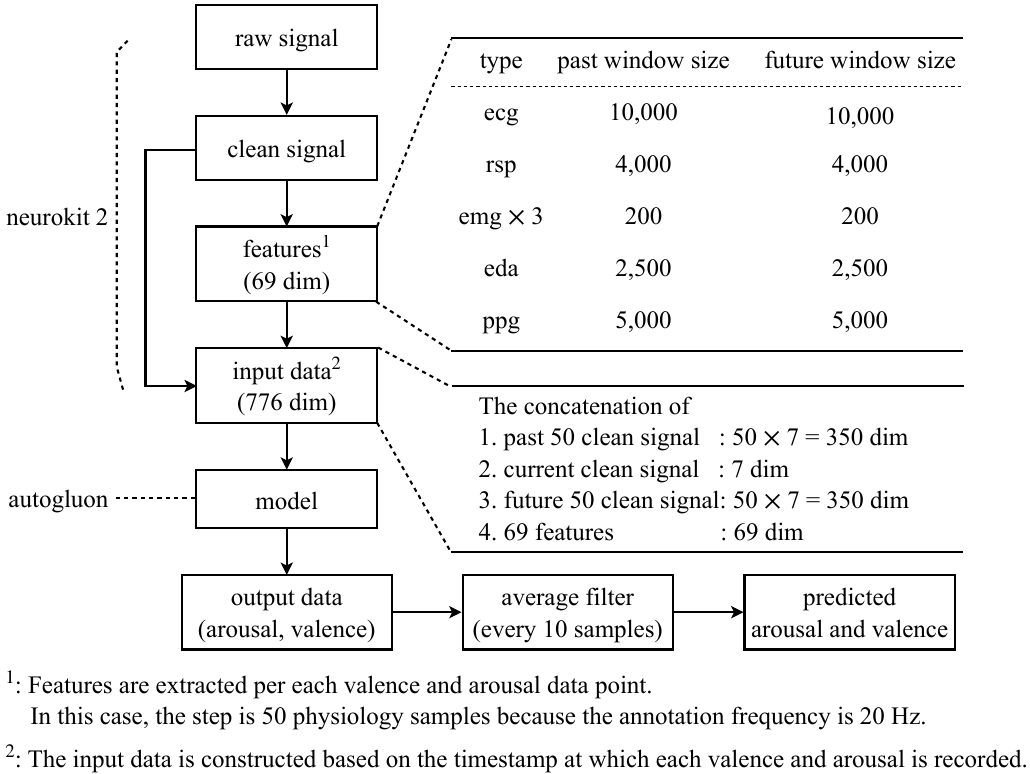}
    \caption{The pipeline used to process each signal, and the post-processing after the machine learning model. Window sizes are represented in samples at 1kHz.}
    \label{fig:flowchart}
\end{figure}

\subsection{Preprocessing}
We used NeuroKit2 to preprocess the physiological signals and feature extraction~\cite{Makowski2021neurokit}. In the case of EMG, we opted to write a custom preprocessing pipeline based on~\cite{perusquia-hernandezSmileActionUnit2021a} in combination with Neurokit.

The EMG pipeline to clean the signal consisted of a series of notch filters at frequencies of 60, 120, 180, and 240~Hz with a notch width of 3~Hz, followed by a bandpass filter with cutoff frequencies of 5~Hz and 250~Hz, and a detrending step. A z-transform is applied to the clean signal, before calculating the root mean square (RMS) over windows of 100~ms. To further smoothen the envelope, a Savitzky-Golay filter of third order with a length of 1~s was applied. This cleaned signal was then input to Neurokit's amplitude function to keep the returned data structure consistent for Neurokit's analyze function. The Neurokit's analyze function was used to extract the features listed in~\autoref{tab:feature_list}. 

Feature extraction was performed over the whole signal using windows of different sizes as shown in~\autoref{fig:flowchart}. These windows are described in samples at 1~KHz. ECG and PPG's window size is longer to sample heart-rate variability. A similar reason applies to respiration. Regarding EDA, a medium-sized window is recommended to decompose the signal into tonic and phasic components. In contrast, the EMG window is shorter, because changes in facial expressions' EMG can happen in the order of milliseconds. As a curious fact, we found a heart-rate artifact in the trapezius EMG. This artifact might be misinterpreted as an EMG feature, but we decided to leave it in because we did not formally distinguish between data types when modeling the data.

\begin{table}[b]
    \centering
    \caption{Feature List}
    \label{tab:feature_list}
    \begin{tabular}{cp{0.7\columnwidth}}

    \toprule
       Signal & Features \\
    \midrule
    PPG (10 dim) & \textbf{Rate}: baseline, max, min, mean, SD, max time, min time, trend linear, trend quadratic, trend R2; \\
    \midrule
    \multirow{3}{*}{ECG (15 dim)} & \textbf{Rate}: baseline, max, min, mean, SD, max time, min time, trend linear, trend quadratic, trend R2; \\
    & \textbf{Phase}: atrial, completion atrial, ventricular, completion ventricular; quality mean; \\
    \midrule
    \multirow{3}{*}{RSP (20 dim)} & \textbf{Rate}: baseline, max, min, mean, SD, max time, min time, trend linear, trend quadratic, trend R2 \\ & \textbf{Amplitude}: baseline, max, min, meanraw, mean, SD;\\
    & \textbf{Phase}, phase completion, RVT baseline, RVT mean; \\
    \midrule
    EDA (6 dim) & peak amplitude, SCR, SCR peak amplitude, SCR peak amplitude time, SCR RiseTime, SCR RecoveryTime; \\
    \midrule
    EMG $\times$ 3 & Activation, Amplitude Mean, Amplitude Max, \\

    (6 $\times$ 3 dim) & Amplitude SD, Amplitude Max Time, Bursts \\
    \bottomrule
    \end{tabular}
\end{table}

The feature vectors were sampled at 20~Hz to match the sampling frequency of the annotations. Additionally, the preprocessed (clean) data temporally surrounding each annotation datapoint were flattened and input as additional features to the modeling block.

\subsection{Model training}
We used a consistent architecture across all four scenarios, but adopted unique strategies for designing the input of model training and generating predictions to accommodate the distinct requirements of various scenarios. 
 
For the architecture, we employed AutoGluon, an open-source AutoML framework developed by AWS, to train our model~\cite{erickson_autogluon-tabular_2020}. This framework expedites the development of machine learning models by automating model training, hyperparameter optimization, and model selection and ensembling. The AutoGluon-Tabular fits a total of 11 models that includes gradient boosting methods (CatBoost, LightGBM, LightGBMLarge, LightGBMXT, XGBoost), extra trees (ExtraTreesMSE), K-nearest neighbors algorithm (KNeighborsDist, KNeighborsUnif), neural networks (NeuralNetFastAI, NeuralNetTorch), and random forests (RandomForestMSE). Furthermore, a weighted ensemble model (WeightedEnsemble\_L2) is fitted and employed to combine the previously-trained models for generating predictions. To achieve optimal performance with AutoGluon, we designate the parameter presets as `best\_quality', allowing AutoGluon to automatically construct robust model ensembles while allocating sufficient training time.

\subsubsection{Across-time scenario}
We aimed to capture the unique characteristics and nuances of each subject's emotional responses and the specific stimuli embedded within the emotional context. Therefore, we trained the model on discrete datasets originating per participant and per video. In this scenario, both training and test sets comprise 240 subsets. The training-test pairs were all collected from the same pool of participants and emotion elicitors that exhibit a one-to-one correspondence. We trained 240 models on each training dataset and subsequently selected the corresponding model to yield predictions on the test sets.

\subsubsection{Across-subject scenario}
The substantial inter-subject variability in physiological responses to stimuli and the inherent limitations of self-report labels, such as subjectivity, bias, and emotional granularity, presents a considerable challenge in developing one-size-fits-all-subjects effective models for affective computing. To generate plausible predictions, we assume that a single video should elicit similar emotions in most participants. Guided by this assumption, we trained a dedicated model for each video. In this scenario, every fold consists of 24 subjects in the training dataset and six subjects in the test dataset, with all participants having watched the same eight videos. We combined data from different subjects of each video as input when training the model. In total, we trained eight models and employed the respective models for affective state estimation when generating predictions for corresponding test set files.

\subsubsection{Across-elicitor scenario}
\begin{figure}[t]
    \centering
    \includegraphics[width=0.8\columnwidth]{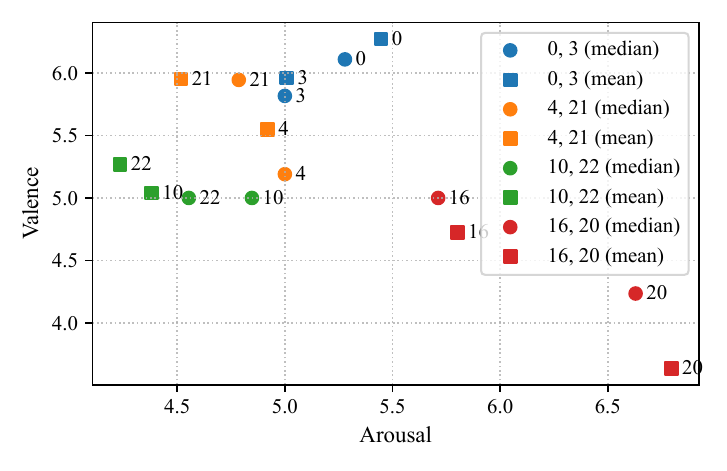}
    \caption{The meta-analysis for determining the quadrant affiliation of videos by the mean and median of video ratings.}
    \label{fig:meta_analysis}
\end{figure}

In this scenario, using only the data from the three quadrants available for estimating arousal and valence could compromise our ability to predict affective states associated with the missing quadrant accurately. To mitigate this concern, we performed a meta-analysis of the training data. We first calculated the mean value of user ratings per video file in the training set to categorize videos into the four affect grid quadrants systematically. By doing so, we could make well-founded assumptions regarding each video's quadrant affiliation. 

Within this scenario, a total of eight videos were provided. By analyzing the composition of the test dataset across four-folds, we categorized the eight videos into four groups: (0, 3), (4, 21), (10, 22), and (16, 20), see~\autoref{fig:meta_analysis}. Among these ratings, it is evident that videos (0, 3) and (16, 20) belong to the high valence high arousal (HV, HA) and low valence high arousal (LV, HA) quadrants, respectively. The categorization of the other two groups is less apparent; therefore, our hypothesis relies on the video with the more prominent rating within each of the two video groups. Consequently, we assumed that video (0, 3) belongs to the (HV, HA) quadrant, (16, 20) belongs to the (LV, HA) quadrant, (10, 22) belongs to the (LV, LA) quadrant, and (4, 21) belongs to the (HV, LA) quadrant. Based on this assumption, we employed only two relevant quadrants to achieve sample balance and maximize the variance along the valence and arousal axes. For instance, when the videos in the test set belong to the (HV, HA) quadrant, we train the valence predictor on the dataset comprising videos from the (LV, LA) and (HV, LA) quadrants, and the arousal predictor on the dataset containing videos from the (LV, HA) and (LV, LA) quadrants. This approach effectively minimizes input bias and ensures more accurate emotional state estimations for the missing quadrant.

\subsubsection{Across-version scenario}
Given that instances of all the affective quadrants are available, albeit in only one version, we aimed to develop a general model that yields robust results by harnessing collective intelligence by applying late fusion. In other words, we assumed there is a ``wisdom of crowds'' effect when combining multiple weak classifiers into one. To this aim, we developed four separate models, each trained on a distinct set of four videos from the training dataset. During the testing process, we input the preprocessed and feature-extracted data from the test set into each of these four models to generate predictions. Then we applied a late fusion strategy to obtain the final estimation. The four predictions from each model were fused by calculating their mean predicted rating values.

\section{Validation results}
\begin{figure}[t]
    \centering
    \includegraphics[width=0.7\columnwidth]{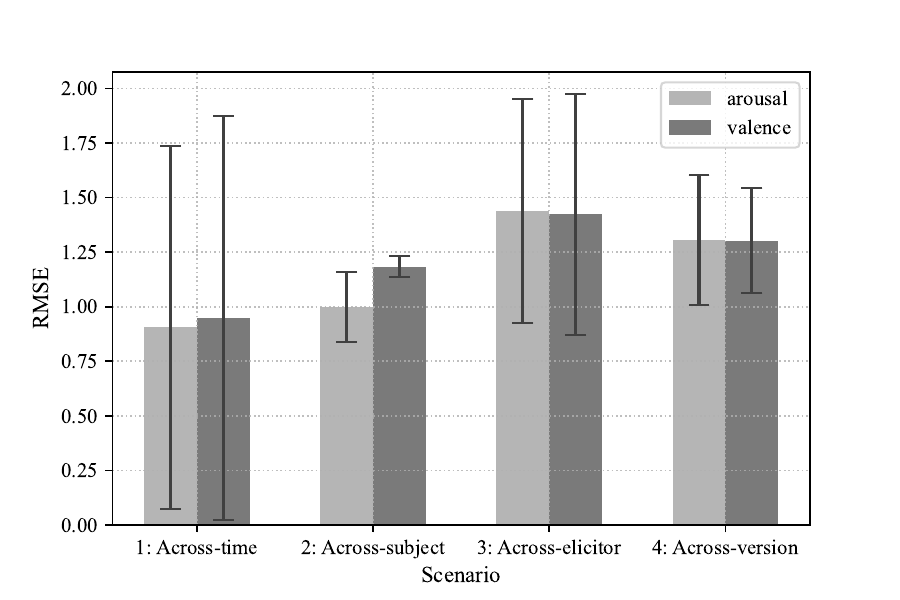}
    \caption{Scenarios-level RMSE. Error bars represent standard deviation.}
    \label{fig:scenarios_level}
\end{figure}

\begin{figure}[t]
    \centering
    \includegraphics[width=0.8\columnwidth]{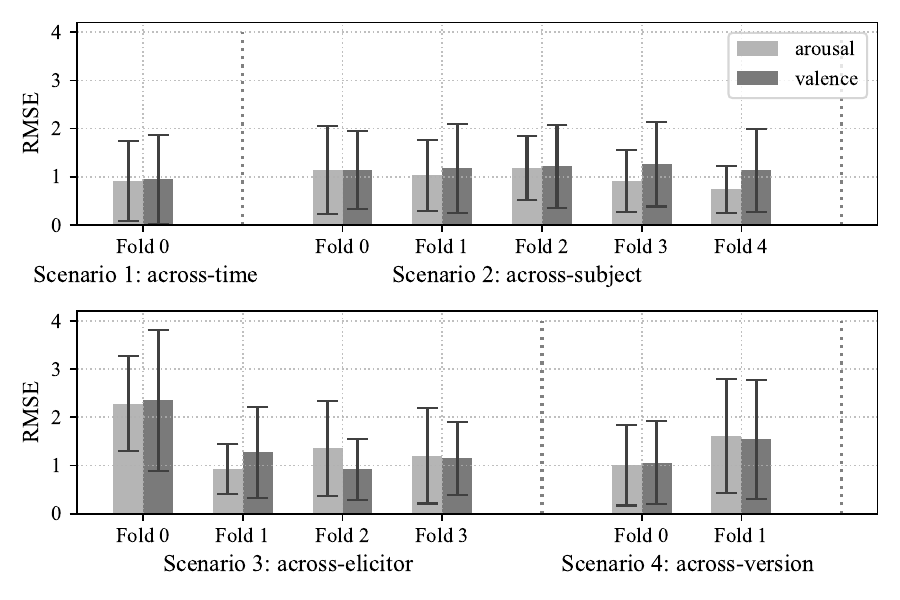}
    \caption{Folds-level RMSE. Error bars represent standard deviation.}
    \label{fig:folds_level}
\end{figure}

\begin{table}[b]
    \centering
    \caption{Folds-level and scenarios-level RMSE}
    \label{table:rmse}
    \begin{tabular}{cccccc}
    \toprule
    \multirow{2}{*}{Scenario} & \multirow{2}{*}{Fold} & \multicolumn{2}{c}{Arousal} & \multicolumn{2}{c}{Valence} \\
        &   &  RMSE & STD & RMSE & STD \\
    \midrule
    Across-time & 0 & 0.91 & 0.83 & 0.95 & 0.93 \\
    \midrule
    \multirow{5}{*}{Across-subject} & 0 & 1.14 & 0.91 & 1.14 & 0.81 \\
        & 1 & 1.03 & 0.74 & 1.17 & 0.92 \\
        & 2 & 1.18 & 0.67 & 1.21 & 0.86 \\
        & 3 & 0.92 & 0.64 & 1.26 & 0.87 \\
        & 4 & 0.74 & 0.48 & 1.13 & 0.86 \\
    \multicolumn{2}{l}{Scenario level} & 1.00 & 0.16  & 1.18 & 0.05 \\
    \midrule
    \multirow{4}{*}{Across-elicitor} & 0 & 2.29 & 0.98 & 2.36 & 1.46 \\
        & 1 & 0.92 & 0.52 & 1.27 & 0.94 \\
        & 2 & 1.35 & 0.99 & 0.92 & 0.64 \\
        & 3 & 1.20 & 0.99 & 1.15 & 0.76 \\
    \multicolumn{2}{l}{Scenario level} & 1.44 & 0.51 & 1.42 & 0.55 \\
    \midrule
    \multirow{2}{*}{Across-elicitor} & 0 & 1.00 & 0.84 & 1.06 & 0.86 \\
        & 1 & 1.60 & 1.18 & 1.54 & 1.23 \\
    \multicolumn{2}{l}{Scenario level} & 1.30 & 0.30 & 1.30 & 0.24 \\
    \bottomrule
    \end{tabular}
\end{table}
The models were assessed using the root mean square error (RMSE) metric. A lower RMSE is better. It has the same units as the valence and arousal annotations.
The final score for the EPiC challenge for our team was 1.19.
Additionally, we report the detailed performance for each scenario on the test set, as reported by the workshop organizers. Our training was done on the full train set to maximize data availability.
For each test data, i.e., the data of one subject and one video, the RMSE is calculated for arousal and valence, respectively.
Then in each fold, the performance is assessed by averaging the RMSE values among all test data.
Similarly, the model performance in each scenario is evaluated by averaging all RMSE values within the scenario.
The final RMSE result was obtained by calculating the mean score on all scenarios and two prediction targets, i.e., arousal and valence.
The scenarios-level RMSE and folds-level RMSE are shown in \autoref{fig:scenarios_level}, \autoref{fig:folds_level}, and \autoref{table:rmse}.

\subsection{Across-time scenario}
The RMSE of predicted arousal and valence are 0.91 and 0.95, with a standard deviation of 0.83 and 0.93, respectively.
Among the 240 test data, the lowest RMSE of arousal and valence is 0 and 0, and the highest is 3.87 and 4.33.
For the predicted results of arousal and valence, the RMSE of 69\% and 66\% of the test results were below 1.
Overall, the prediction error for arousal is slightly lower than that for valence.

\subsection{Across-subject scenario}
The RMSE for the predicted arousal and valence are 1 and 1.1, respectively, with standard deviations of 0.16 and 0.04 across the five folds.
In each fold, the predicted arousal RMSE is consistently lower than the predicted valence RMSE.
This difference is especially noticeable in fold 4, where the RMSE and standard deviation for predicted arousal are 0.74 and 0.48, respectively, in contrast to the valence RMSE and standard deviation, which are 1.13 and 0.86, respectively.

Despite the seemingly positive results in this scenario, there is a limitation on how we predicted the final ratings. We assumed that training and testing across elicitors would help to predict better the outcomes in the ratings done by other people not seen in the dataset. However, in the real world, we do not have information about the type of stimuli used. Therefore, the performance will probably be reduced, as exemplified in the following section. 

\subsection{Across-elicitor scenario}
Our model yielded the highest RMSE values, with the RMSE for arousal and valence being 1.44 and 1.42, respectively, and standard deviations of 0.51 and 0.55.
We note that the high RMSE was due to poor prediction results in fold 0, where the RMSE values were twice those observed in other folds, reaching 2.29 and 2.36 for arousal and valence, respectively.
By analyzing the data for this scenario, we noticed that a potential cause for the high RMSE lies in the significant deviation between the test data and the training data in fold 0. The training data does not include similar patterns in the test data.
As shown in~\autoref{fig:videoValenceArousal}, the test data in fold 0 contains videos 16 and 20 in the upper left quadrant of the affective grid. The rating patterns of arousal and valence differ from the data in the other three quadrants, which were used as training data. This suggests that the larger variations in ratings characteristic of negative, high-arousing emotions are not present in the other types of emotion.
Furthermore, these results also demonstrate the reliance of our model on data similarity, indicating a weaker generalization capability for novel data patterns.

\subsection{Across-version scenario}
In this scenario, the prediction RMSE using our model is 1.30 for both arousal and valence, with standard deviations of 0.30 and 0.24, respectively.
Although each of the two groups' data in this scenario covered all four emotional states, the cross-validation results revealed that our model's RMSE in fold 0 was 0.5 lower than in fold 1.
This suggests that an appropriately balanced training set, encompassing all four emotional states, can significantly enhance the model's generalization capabilities.

\section{Revisiting assumptions}

\subsection{Lag between physiological signals and ratings}
When preparing the data to train our models, we assumed that the physiological changes happen at different speeds depending on the measurement metrics used.
Here we investigate the effect of the time delay in reporting emotions.
In particular, we followed the general procedure described by \citet{schmittBorderAcousticsLinguistics2016}. We shifted the features forward in time in steps of $0.005$~s up to a maximum of $0.05$~s and trained a Gated Recurrent Unit (GRU) model to predict arousal and valence. Note that the annotations were performed at $20$~Hz while the physiological signals were sampled at $1000$~Hz. Then, we experimented with using each individual signal as input to the model in isolation before combining all signals as input. 
The analysis results are in \autoref{tab:lag}.
The results suggest that predictive performance generally improves when accounting for annotation delays. However, the delay yielding the most empirical gains varies for each biosignal, and we often found several minimum values. A further investigation of the timing relationships between physiological change, experience, and annotation is needed to understand when the differences significantly disrupt the predictions.


\begin{table}[b]
    \centering
    \caption{RMSE at different rating-biosignal delays. Bold values are the minimum.}
    \label{tab:lag}
    \setlength{\tabcolsep}{4pt}
    \begin{tabular*}{\columnwidth}{@{}l@{\extracolsep{\fill}}ccccccccc@{}}
    \toprule
    \multirow{3}{*}{Delay} & \multicolumn{9}{c}{RMSE ($10^{-3}$)}\\
      &   ALL & BVP & ECG & EMG & EMG & EMG & GSR & RSP & SKT\\
      &    &  &  & $coru$ & $trap$ & $zygo$ & & &\\
    \midrule
0 & \textbf{1.21} & 1.21 & 1.22 & 1.24 & 1.24 & 1.23 & 1.24 & \textbf{1.19} & 1.22 \\
0.005 & 1.24 & 1.22 & 1.22 & 1.21 & 1.24 & 1.22 & 1.21 & 1.20 & 1.20 \\
0.01 & 1.22 & 1.21 & \textbf{1.20} & \textbf{1.20} & 1.23 & 1.25 & 1.21 & 1.27 & 1.21 \\
0.015 & \textbf{1.21} & 1.23 & 1.22 & 1.21 & 1.20 & 1.23 & 1.22 & 1.24 & 1.20 \\
0.02 & 1.24 & 1.23 & 1.22 & 1.23 & 1.23 & 1.24 & 1.2 & 1.21 & 1.20 \\
0.025 & 1.23 & 1.21 & 1.23 & 1.22 & 1.22 & 1.22 & 1.22 & 1.21 & 1.23 \\
0.03 & 1.22 & \textbf{1.20} & 1.21 & 1.22 & 1.24 & \textbf{1.21} & 1.22 & 1.20 & \textbf{1.19} \\
0.035 & \textbf{1.21} & 1.23 & 1.21 & 1.26 & \textbf{1.19} & 1.23 & \textbf{1.18} & 1.20 & 1.24 \\
0.04 & 1.23 & 1.23 & 1.23 & 1.20 & \textbf{1.20} & 1.24 & 1.23 & 1.23 & 1.24 \\
0.045 & 1.23 & 1.21 & \textbf{1.20} & \textbf{1.20} & 1.22 & 1.23 & 1.22 & 1.21 & 1.23 \\
0.05 & 1.26 & 1.21 & 1.23 & 1.23 & 1.23 & 1.23 & 1.21 & 1.22 & \textbf{1.19} \\
    \bottomrule
    \end{tabular*}
\end{table}

\subsection{The gradual nature of changes in emotion}
Qualitatively, we noticed that the predictions from our models were characterized by more high-frequency changes than the annotations provided in the training data, which change more gradually over time. Consequently, we utilized a moving average window comprising 10 samples, equivalent to a 2~Hz low-pass filter.
The choice of the window length follows from the observation that the annotations were provided at $20$~Hz, so the Nyquist frequency is $10$~Hz -- only changes at 10~Hz can be measured with the joystick described in the dataset. Furthermore, we assumed people would not make more than two abrupt changes per second. Further investigation is required to establish whether the nature of emotion changes is a gradual process, or whether the relative smoothness of the ground-truth ratings is an annotation artifact.  

\begin{figure*}[t]
    \centering
    \includegraphics[width=0.82\textwidth]{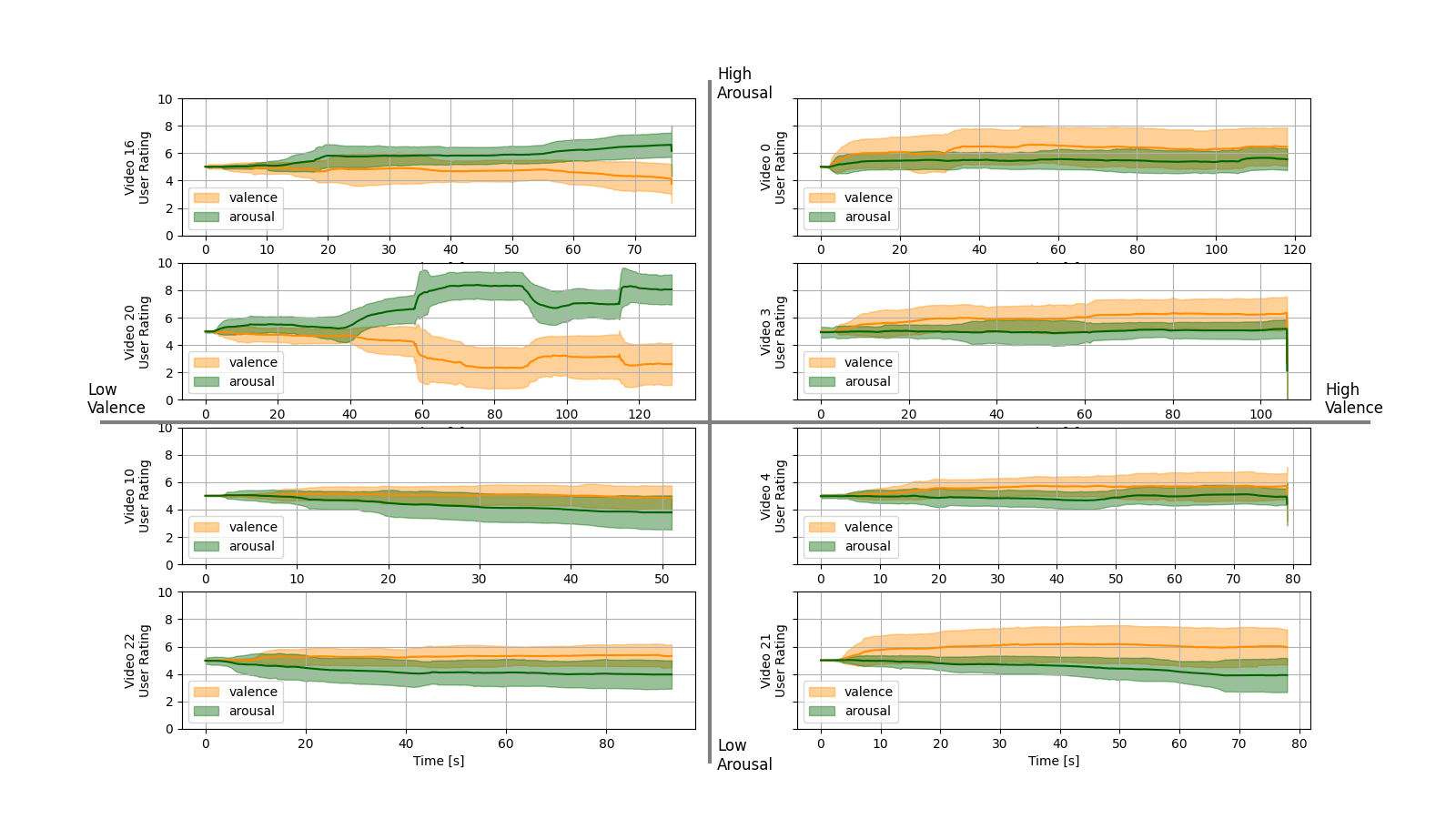}
    \caption{Plot of the rating averages per each stimuli video. Shaded areas represent the standard deviation among participants.}
    \label{fig:videoValenceArousal}
\end{figure*}

\subsection{Single- and multi-label predictors}
\begin{figure}[t]
    \centering
    \includegraphics[width=0.7\columnwidth]{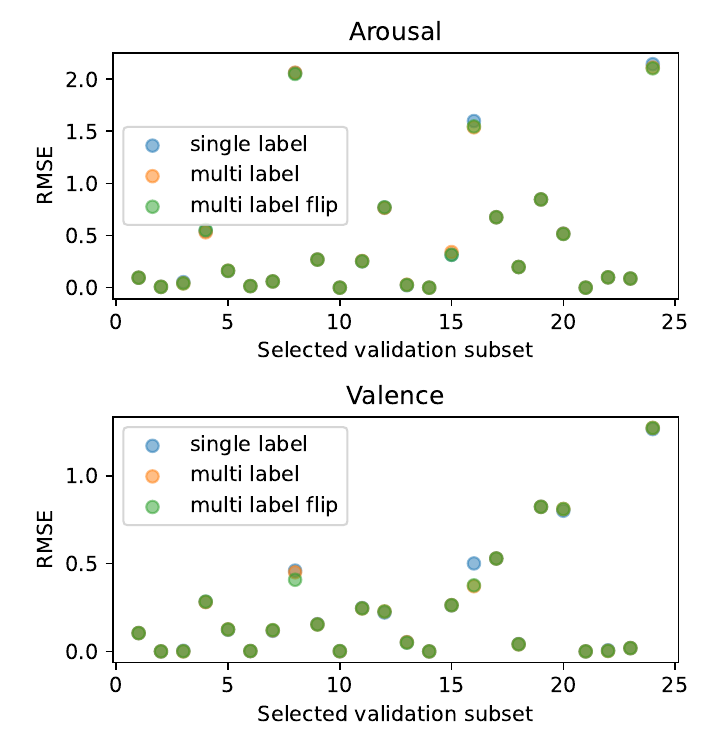}
    \caption{The comparison between single and multi-label predictors. The ``single label'' represents independent prediction of valence and arousal. ``Multi label'' and ``multi label flip'' correspond to predicting arousal after predicting valence and vice versa.}
    \label{fig:single_and_multi_label}
\end{figure}

Considering that participants simultaneously rated their emotions for valence (X-axis) and arousal (Y-axis) using a two-dimensional joystick, we speculated on the potential connection between these two values, despite the orthogonal nature of the valence-arousal model's axes.
To evaluate this hypothesis, we extracted 24 datasets from scenario 1, each containing data from the combination of six participants and four videos. We compared the following approaches: (1) independent prediction of valence and arousal, (2) predicting valence first and then using both physiological signals and predicted valence values for arousal prediction, and (3) predicting arousal first and subsequently using the predicted arousal values for valence prediction. This comparison investigated any potential connections between valence and arousal predictions.
As demonstrated in~\autoref{fig:single_and_multi_label}, the performance differences among these three strategies are minimal. As a result, owing to the slower training process associated with multi-label predictors compared to single-label cases, we ultimately chose to employ the independent prediction strategy for this competition.

\section{Discussion and future directions}
We introduced an attempt to address the EPiC Challenge. We predicted continuous valence and arousal ratings from several biosignals, across four scenarios. We used readily available algorithms, with our novelty being (a) the window choices for feature calculation; (b) the use of data around each annotation point; and (c) our use of theoretical assumptions to maximize data variance in each scenario's training. Our overall RMSE for the four validation scenarios was 1.19, as provided by the competition organizers. This result still has considerable room for improvement compared to previous work on predicting continuous valence and arousal annotations from physiological data, and can be used as a baseline for future regression studies on the CASE dataset.
As expected from the literature, fitting a personalized model to predict a single persons's reaction at a future point (across-time) is an easier problem than in the other scenarios. To extend our model to other, unseen people, we used information about the stimuli, and capitalized in our knowledge of the affective context in which the data was collected. By training several models per stimuli, we reduced the RMSE in the across-subject scenario. However, this strategy is unlikely to work in the real world, as we typically would not have information about the stimulus. A similar approach was used in the across-elicitor scenario. By examining the results~(\autoref{fig:videoValenceArousal}), we hypothesize that high arousal and low valence emotions display abrupt physiological changes not present in other affective quadrants. Future work should explore whether this is consistently true, and devise a method to tackle the lack of information during training. Furthermore, the results of the across-version validation signal that expected affective messages depicted in a stimulus might not produce the same effect in different individuals. Future work should validate if this is the case, and assess if a weighted late fusion provides improvements with respect to an averaging function. This would also validate or refute whether our bet for a ``wisdom of a crowd classifier'' is suitable. Finally, future work should be done to formally assess if end-to-end methods outperform ensemble methods and feature engineering similar to those used in this work.

\section*{Ethical Impact Statement}
This research was a data analysis of the CASE dataset~\cite{sharmaDatasetContinuousAffect2019a}. All data provided is anonymous, and obtained following the Declaration of Helsinki.  
Our results have several limitations. The sample size is only 30 people, and no mentions of the cultural background of the participants are made in the dataset. The interpretation of the stimuli, and therefore the ratings and physiological changes, might differ from person to person. Therefore, our results need to be replicated in other corpora.
Moreover, we built our model based on certain assumptions that are described, but not formally validated. These assumptions might not yield the same performance in other datasets. Finally, our results are biased to better predict the situations in the dataset.
Therefore, our model should be used with caution.


\printbibliography
\end{document}